\documentclass[]{article}
\usepackage{proceed2e}

\usepackage{graphicx} 
\usepackage{subfigure} 

\usepackage{algorithm}
\usepackage{algorithmic}

\usepackage{amsmath, amssymb}

\author{ {\bf Evan Ettinger and Yoav Freund} \\ Computer Science and Engineering Dept. \\ UC San Diego \\ \{eettinger, yfreund\}@cs.ucsd.edu}

\title{Particle Filtering on the Audio Localization Manifold}

\begin{document} 

\maketitle

\begin{abstract} 
We present a novel particle filtering algorithm for tracking a moving sound source using a microphone array.  If there are N microphones in the array, we track all $N \choose 2$ delays with a single particle filter over time.  Since it is known that tracking in high dimensions is rife with difficulties, we instead integrate into our particle filter a model of the low dimensional manifold that these delays lie on.  Our manifold model is based off of work on modeling low dimensional manifolds via random projection trees~\cite{Dasgupta09}.  In addition, we also introduce a new weighting scheme to our particle filtering algorithm based on recent advancements in online learning.  We show that our novel TDOA tracking algorithm that integrates a manifold model can greatly outperform standard particle filters on this audio tracking task.
\end{abstract} 

\section{Introduction}
\label{intro}
There is an increasing interest in locating audio sources with a microphone array as a means to direct the pointing of a camera.  Camera pointing applications include video conferencing, surveillance, game playing and interactive displays.  In addition, speech enhancement with microphone arrays rely critically on knowing the correct source location.

One popular method for locating an audio source is based on measuring the delays observed between spatially separated pairs of microphones known as the \emph{time delay of arrival} (TDOA).  For locating a source a two stage process can be employed: First TDOAs for all pairs of microphones are estimated, and then a source location is derived from this delay information.  If microphone positions are given, the second step becomes approximately solving a set of non-linear physical equations such as in \cite{Dibiase01}.  However, localizing an audio source accurately in a large room requires that the microphones are far apart from each other.  As a result of placing the microphones far apart, it becomes difficult to estimate their positions within a coordinate system accurately.  If the positions are not known, then a regressor can be learned that maps TDOAs to camera pointing directives as in \cite{Ettinger08, Cheamanunkul09}.

In this work we focus on accurately estimating and tracking TDOAs for a microphone array in a large room.  There is an extensive literature on using particle filters for tracking audio sources when the microphone positions are known \cite{Ward03, Lehmann07}.  Since positional information is known, the state space for the particles is typically only two or three spatial dimensions for the location of the sound source.  When the microphone positions are not known and we attempt to track in the native TDOA space we become victim to the slew of problems that come with tracking in high dimensions.  With N microphones in the array each pair has a TDOA that needs to be tracked making the state space be of dimension $D = {N \choose 2}$.  $D$ can be quite large for a microphone array in a large room.

To alleviate the problem of high dimensionality we propose an addition to the particle filter that includes a restriction on the state space of particles to that of a low dimensional manifold.  Underlying the $D$ dimensions of a TDOA measurement are only three degrees of spatial freedom for the sound source to move in.  Each 3-d spatial location creates a unique TDOA vector which varies smoothly with smooth variations in the spatial location.  We model this low dimensional manifold using a tree-based spatial partitioning data structure combined with principal components analysis.  Our tree structure is based on work on random projection trees, which have been shown to adapt to low dimensional intrinsic structure when the data itself lies in a high dimensional space \cite{Dasgupta09}.

We also investigate in this work a new particle filter based on work from the online learning body of literature.  In particular we focus on work from combining expert advice via the normal hedge algorithm~\cite{Chaudhuri09}.  For particle filters, each expert is itself a particle that predicts a state at each time step.  The normal hedge particle filter gives both a new particle weighting scheme and a natural resampling scheme for particles based on the fact that the algorithm explicitly gives zero weight to poorly performing particles.  Using normal hedge in the particle filtering framework has been initially explored in~\cite{Chaudhuri-2-09}.  This is the first time this algorithm has been applied to the TDOA tracking problem, and to the best of our knowledge, any practical problem to date.

The rest of the paper is organized as follows.  Section~\ref{tdoa} briefly discusses how we estimate TDOAs for a given pair of microphones via the phase transform.  Section~\ref{rptrees} discusses random projection trees and how they adapt to low dimensional intrinsic structure.  Section~\ref{normal} discusses our particle filter implementation that includes the model of the manifold.  Finally in section~\ref{experiments} we discuss some experiments on tracking TDOA vectors with real-world data collected from an interactive display.

\section{Time Delay of Arrival}
\label{tdoa}
One very popular method for estimating a TDOA given frames of audio from a pair of microphones is to use a generalized correlation technique such as the phase transform otherwise known as PHAT \cite{Ward03}.  PHAT is a normalized cross correlation technique that removes the magnitudes of the amplitude information from the audio signals putting the emphasize on aligning the phase components.  Define $R^{p}(\tau)$ as the PHAT correlation between microphone pair $p$ at time delay $\tau$, then the TDOA is often estimated by
\begin{equation}
  \hat{\Delta}^{p} = \arg\max_{\tau} R^{p}(\tau) \label{eqn:tdoacalc}
\end{equation}
However, in a reverberant environment there are often spurious peaks in $R^{p}$ from either line noise or multipath reflections.  In these cases the true TDOA may not be the largest peak in the PHAT correlation.  By using particle filters we are able to leverage this secondary peak information when formulating a likelihood function that incorporates the entirety of the observation $R^{p}$.  This gives the particle filtering method a robustness over traditional approaches that depend on the accuracy of Equation~(\ref{eqn:tdoacalc}) over all pairs $p$.

\section{Modeling the Manifold}
\label{rptrees}

\begin{figure}[t!]
\centering
\includegraphics[width=0.9\columnwidth]{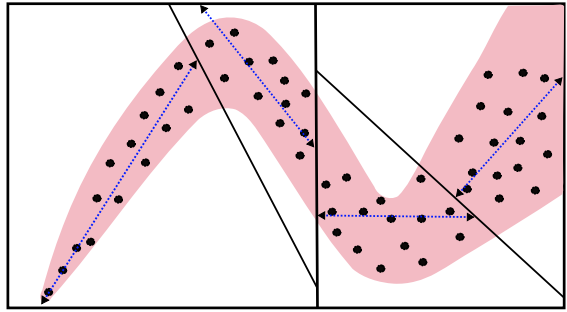}
\caption{A toy dataset whose distribution is shown in pink. A PDTree of height one is built with top principal direction shown in each leaf node.\label{fig:PDTree}}
\end{figure}

A TDOA vector has only three underlying spatial degrees of freedom.  If the microphone positions were known, then the physics equation for the TDOA between microphone pair $p=(i,j)$ is
\begin{equation}
  \Delta^{p} = \frac{\|m_i - s\|_2 - \|m_j - s\|_2}{c} \label{eqn:tdaodef}
\end{equation}
where $m_i$ is the position of microphone i, $s$ is the source location and $c$ is the speed of sound in air.  In this work we assume no such knowledge of $m_i$, but nevertheless the same physical principals apply.  As $s$ varies smoothly, so does $\Delta^{p}$.  So even though the vector containing the TDOAs for all microphone pairs has $D$ components, the real underlying dimensionality is only three.  We call this lower dimensional smooth structure the \emph{TDOA manifold}.

Modeling the TDOA manifold for a particular array configuration is an integral part of the particle filtering algorithm we present in Section~\ref{normal}.  Our model is based off of the random projection tree spatial partitioning algorithm whose details can be found in \cite{Dasgupta09}.  A random projection tree (RP-tree) is a binary tree that recursively splits a dataset into two subsets.  It is constructed in nearly the same way as a KD-tree but instead of recursively splitting the dataset along a single coordinate axis, the data is first projected onto a random direction and then split near the median of these projections.  RP-trees have been effectively used  as a means for vector quantization and for regression problems when the data has much lower intrinsic dimensionality than it's ambient dimension\cite{Samory09, Freund07}.  

The intrinsic dimensionality of a dataset can be measured in a variety of ways including Assouad dimension or fraction of variance explained by a PCA at the appropriate neighborhood size.  RP-trees guarantee that if the data falling in a given node $n$ of the tree has intrinsic dimensionality $d$, then all cells $O(d)$ levels below $n$ have at most half the data diameter.  This guarantee depends only on the intrinsic dimensionality of the data $d$ and not the ambient dimensionality $D$.  Therefore, we can expect a rapid convergence to the manifold structure from such a partitioning tree.

To model the TDOA manifold we first collect a training set of TDOA vectors sampled from the room containing our fixed microphone array.  This can be done by using a white noise source and moving it throughout the room.  Since white noise is random, the TDOAs measured via PHAT using Equation~\ref{eqn:tdoacalc} are very reliable training data after some simple outlier removal.  Another way to collect such a training set is from interactions by people with an interactive display as in~\cite{Cheamanunkul09}.  

The tree we build in this work is similar to an RP-tree but uses principal components analysis instead of random projections.  We call this tree a PD-tree and it has been shown empirically that these trees adapt to intrinsic dimensionality well in practice~\cite{Verma09}.  A PD-tree recursively partitions the training set by projecting the data onto its top principal direction and then choosing the median of these projections to be the splitting point.  A depiction of a PD-tree of height 1 on a toy dataset is given in Figure~\ref{fig:PDTree}.  We find that in practice using the top principal direction lends to quicker convergence to the underlying manifold compared to using random directions.

At each node of the PD-tree we store the mean and top k principal directions of the data that belongs to the node.  We use this tree as a means of denoising TDOAs.  For a given TDOA vector, find the corresponding leaf node it belongs to and then project it onto the affine space spanned by the top k eigenvectors stored in that leaf node.  This is effectively a projection onto the manifold where the manifold is modeled piecewise by PCAs of local neighborhoods.

\section{Particle Filters \& Normal Hedge}
\label{normal}
In this section we briefly describe a standard particle filtering algorithm as it relates to the TDOA tracking problem.  We then introduce a new particle filtering algorithm with a new weighting and particle resampling scheme based on results from online learning. 

\subsection{Particle Filtering Framework}
Particle filtering is an approximation technique used to solve the Bayesian filtering problem for state space tracking first proposed in~\cite{Gordon93}.  For TDOA tracking, the state space $X_t$ is composed of each of the $D$ time delays.  A weighting over $m$ particles is chosen to approximate the posterior density at time $t$ over this state space.  A good tutorial discussing particle filtering and its many variants can be found in~\cite{arulampalam2002}.

\renewcommand{\algorithmicrequire}{\textbf{Initial Assumptions:}}

\begin{algorithm}[tb]
   \caption{SIR Particle Filter for TDOA Tracking}
   \label{alg:pf}
\begin{algorithmic}[1]
   \REQUIRE At time t-1, we have the following:
   \begin{enumerate}
     \item Set of $m$ particles $X_{t-1}^i$ for $i \in \{1,\dots,m\}$.
     \item A collection of PHAT correlation observations at time t $R_t(\tau)$ for each pair of microphones.
     \item Each particle's weight $w_{t-1}^i$, a discrete representation of the posterior $Pr(X_{t-1} | R_{1:t-1})$.
     \item A likelihood function $\mathcal{L}(R_t, X_t) \propto Pr(R_t | X_t)$.
     \item A resampling variance parameter $\Sigma_r$
   \end{enumerate}
   \STATE {\bfseries Resampling:} Resample $m$ new particles and add independent Gaussian noise 
   \begin{equation*}
     X_t^i = \tilde{X}_t^i + n_i
   \end{equation*}
   where $\tilde{X}_t^i$ is drawn according to $\{w_{t-1}\}$ from the set of particles at $t-1$ and $n_i \sim N(0,\Sigma_r)$.
   \STATE {\bfseries Weight Update:} Assign each particle a likelihood weight according to 
   \begin{equation*}
     w_t^i = \mathcal{L}(R_t^p, X_{t}^i)
   \end{equation*}
   Normalize weights so that they sum to 1.
   \STATE {\bfseries Prediction:} Predict state according to the weighted average
   \begin{equation*}
     \sum_{i=1}^m w_t^i X_t^i
   \end{equation*}
\end{algorithmic}
\end{algorithm}

One popular variant is the sampling importance resampling (SIR) particle filter.  We examine this filter for our purposes since it has been shown to work well for audio tracking when a coordinate system is known and can be used for the state representation~\cite{Ward03, Lehmann07}.  A single iteration of such a SIR particle filtering algorithm for the TDOA tracking problem is given in Algorithm~\ref{alg:pf}.  At each time step the algorithm goes through a resampling, a prediction and an update stage.  The key decisions for optimizing the performance of this TDOA tracking algorithm are:
\begin{enumerate}
   \item The choice of $\mathcal{L}(R_t,X_t)$, the likelihood function of the observation given the state.  For a given state $X_t$ the likelihood function measures how likely it is to have observed the PHAT correlation $R_t$.  This function should be chosen so that the likelihood function is largest when the coordinates of state $X_t$ is nearby many of the peaks in each of the corresponding $R_t$.  However, modeling the true likelihood of the PHAT observation given the state is problematic since it is affected by issues such as line noise and multipath reflections.  This makes accurately modeling this likelihood rather challenging, and instead a pseudo-likelihood is employed.
   \item The total number of particles $m$.  The larger $m$ is the more computational load the system must undertake.  Minimizing $m$ while not sacrificing performance is of paramount importance for real time implementations.
   \item The covariance of the resampling noise, $\Sigma_r$.  We assume a very simple model for the state space in what follows, namely that sound sources do not move too quickly. We should choose the size of $\Sigma_r$ to match how quickly we expect sound sources to be moving.  More expressive state spaces that take into account the velocity or higher order moments of each TDOA coordinate are not explored in this work. 
\end{enumerate}

We integrate the manifold modeling discussed in the previous section at the resampling stage.  That is, after resampling a new particle it can be denoised by projecting it through the trained tree model.  This will disallow particles to drift off into regions where TDOAs can not be created by true sound sources.

\subsection{Normal Hedge Particle Filtering}

To discuss the differences between the SIR particle filter and the normal hedge version we must first introduce some terminology from the online learning body of literature.  Normal hedge is an online learning algorithm that attempts to learn how to combine predictions from experts at each time step so as to compete with the predictions of the best set of experts in the collection.  

The algorithm maintains a distribution over the experts $w_t^i$.  At each time step each expert suffers a bounded loss $\ell^i_t$ which is a function of the observation and the experts prediction at time $t$, typically squared, absolute or log-loss.  Finally, the algorithm suffers the loss $\sum_i w_t^i \ell_t^i$.  The cumulative loss at time $t$ for expert $i$ is then $L^i_t = \sum_{i=1}^t \ell_t^i$ (cumulative loss for the algorithm, $L^{A}_t$ is similar).  Often the goal of an online learning algorithm is to maintain a distribution such that $L^{A}_t$ is small relative to that of the best expert in the set, $\min_i L_t^i$.  Instead of competing with the best expert in hindsight, normal hedge attempts to compete with the top $\epsilon$-quantile of $L_t^i$.  This setting is useful when the number of experts is very large and it is expected that many of the experts will perform very similarly.  

A key concept in online learning is the \emph{regret at time t} of the algorithm $R^{A}_t = L^A_t - L^i_t$ to a particular expert $i$.   The theoretical guarantee of normal hedge is that the algorithm's regret at time $t$ to the $\lfloor \epsilon N \rfloor$-best expert is small.  This is not as strong as the regret to the \emph{best} expert in hindsight being small, but is very applicable when an $\epsilon$ fraction of experts in fact predict well.  We will exploit this fact in our tracking problem.  In addition, unlike many other online learning algorithms which have a learning rate parameter that controls how aggressive the $w_t^i$ updates are made, normal hedge has no such parameter to tune.  A detailed explanation of normal hedge in the online setting can be found in~\cite{Chaudhuri09}.

\begin{algorithm}[tb]
   \caption{NH Particle Filter for TDOA Tracking}
   \label{alg:nh}
\begin{algorithmic}[1]
   \REQUIRE At time t-1, we have the following:
   \begin{enumerate}
     \item Set of $m$ particles $X_{t-1}^i$ for $i \in \{1,\dots,m\}$.
     \item A collection of PHAT correlation observations at time t $R_t(\tau)$ for each pair of microphones.
     \item Each particle's weight $w_{t-1}^i$.
     \item A scoring function $\mathcal{L}$ for how well $X_t$ matches the observation $R_t$.
     \item A resampling variance parameter $\Sigma_r$
   \end{enumerate}
   \STATE {\bfseries Weight Update:} Update the discounted cumulative regret of each particle and each particle's weight using~(\ref{eqn:cumloss})--(\ref{eqn:ct}).  Normalize weights so that they sum to 1.
   \STATE {\bfseries Prediction:} Predict the state according to the weighted average.
   \begin{equation*}
     \sum_{i=1}^m w_t^i X_t^i
   \end{equation*}
   \STATE {\bfseries Resampling:} For each particle with \emph{zero weight}, resample a new particle
   \begin{equation*}
     X_t^i = \tilde{X}_t^i + n_i
   \end{equation*}
   where $\tilde{X}_t^i$ is drawn according to $\{w_{t-1}\}$ from the set of particles at $t-1$ and $n_i \sim N(0,\Sigma_r)$.  Also, reassign the cumulative regret to be the same as that of $\tilde{X}_t^i$.
\end{algorithmic}
\end{algorithm}

Normal hedge is easily adapted to the problem of tracking with particle filters.  Here the experts predict a state at each time step, exactly the same as what a particle does in SIR particle filtering.  At each time step the experts suffer a loss which is based on the same likelihood function $\mathcal{L}(R_t,X_t)$ as discussed for particle filters.  Instead of calculating the cumulative loss of each expert, we maintain the \emph{discounted} cumulative regret.

\begin{align}
  G_t^i &= (1-\alpha)G_{t-1}^i + (\mathcal{L}(R_t,X_{t-1}^i) - g_t^{A}) \label{eqn:cumloss} \\
  g_t^{A} &= \sum_{i=1}^m w_{t-1}^i\mathcal{L}(R_t,X_{t-1}^i) \label{eqn:instantgain}
\end{align}
Where $\mathcal{L}$ is the likelihood scoring function used in the generic particle filtering algorithm, $g_{t}^{A}$ is the weighted likelihood of all the particles, and $\alpha$ is the discounting factor.  The second term in~(\ref{eqn:cumloss}) is the instantaneous regret between the algorithm and the $i^{th}$ expert.  The choice of $\alpha$ determines how long the memory is for the discounted cumulative regret, which determines how far back a particle must suffer for mistakes in the past.  Given $G_t^i$ for each particle, we use the normal hedge weighting update to determine each particle's weight.
\begin{equation}
  w_t^i = \frac{[G_t^i]_+}{c_t}\exp{\frac{([G_t^i]_+)^2}{2c_t}} \label{eqn:weight}
\end{equation}
$[A]_+$ denotes $\max(A,0)$ and $c_t$ is the solution to
\begin{equation}
  \frac{1}{m} \sum_{i=1}^m \exp{\frac{([G_t^i]_+)^2}{2c_t}} = e \label{eqn:ct}
\end{equation}
where $e$ is Euler's number.  Note that the weighting is very aggressive since it is doubly exponential in $G_t^i$.  A more in depth discussion of the normal hedge particle filter can be found in~\cite{Chaudhuri-2-09}.  An instantiation of such an algorithm for the TDOA tracking problem is given in Algorithm~\ref{alg:nh}.

There are a few things to note about this algorithm.  First, the resampling scheme for particles is built into the normal hedge framework since particles get assigned zero weight when they have a non-positive discounted cumulative regret.  Therefore, when an iteration occurs where a particle is found to have weight zero, a resampling step is undertaken that replaces it near a particle that currently is performing better than the algorithm's cumulative regret.  This leads to a very natural resampling scheme that undergoes much less sampling per iteration than the SIR particle filter which resamples every particle every iteration.

The second thing to note is that there are no probabilistic assumptions about $\mathcal{L}$.  The only requirement is that the user provide a scoring function, denoted $\mathcal{L}$ by which the particles are judged by, but unlike SIR particle filters it need not be an accurate representation of the true likelihood.  The introduction of a scoring function to which performance can be guaranteed makes for a strong match with practical considerations.

\subsection{Choice of Scoring Function}
What remains to be discussed is how we define our likelihood (scoring) function $\mathcal{L}$.  It is difficult to accurately define the likelihood of an observation of a group of PHAT correlations given a particular state.  Instead, we define a pseudo-likelihood, $\mathcal{L}$.  We'd like $\mathcal{L}$ to be large when the state is near large peaks in the PHAT correlations series.  Moreover, we would like to encourage the particles to track these peaks over time, so they should be attracted in the direction of these peaks as well.  

To identify the peaks in a particular PHAT function we take a simple z-scoring method.  For each PHAT correlation $R_t^p$ let it undergo a z-scoring transform as follows:
\begin{equation}
  Z_t^p(\tau) = \left[ \frac{R_t^p(\tau) - \mu_t^p}{\sigma_t^p} - C \right]_+ \label{eqn:zscore}
\end{equation}
where $\mu_t^p, \ \sigma_t^p$ are the mean and standard deviation of $R_t^p$ over a fixed bounded range of $\tau$, and $C$ is a constant requiring that peaks be at least C standard deviations above the mean.  This performs well to find a fixed small number, $K_t^p$ of peaks in each $R_t^p$ since PHAT sequences typically have a small number of very large peaks relative to the rest of the series.  Now we define a pseudo-likelihood function as follows:
\begin{equation}
  \mathcal{L}(Z_t, X_t) = Z_0 + \sum_{p=1}^{D}\sum_{l=1}^{K_t^p} Z_t^p(\tau_l^p)\mathcal{N}(\tau_l^p; X_t^p, \sigma_z^2) \label{eqn:scoring}
\end{equation}
where $X_t^p$ is the TDOA for pair $p$ for this state, $\mathcal{N}(x;\mu,\sigma^2)$ is the density under a normal distribution evaluated at $x$ with mean $\mu$ and variance $\sigma^2$, and $Z_t^p$ has $K_t^p$ non-zero entries each of which are at $\tau_l^p$.  The parameter $Z_0$ is the background likelihood that determines how much likelihood is given to any state.  The variance parameter $\sigma_z^2$ controls how much weighting is given relative to how far each state is from the peaks in the corresponding PHAT series.  A similar pseudo-likelihood function is given in~\cite{Ward03}.

\begin{figure*}[t]
\vskip 0.2in
\begin{center}
\centerline{
\includegraphics[width=\columnwidth]{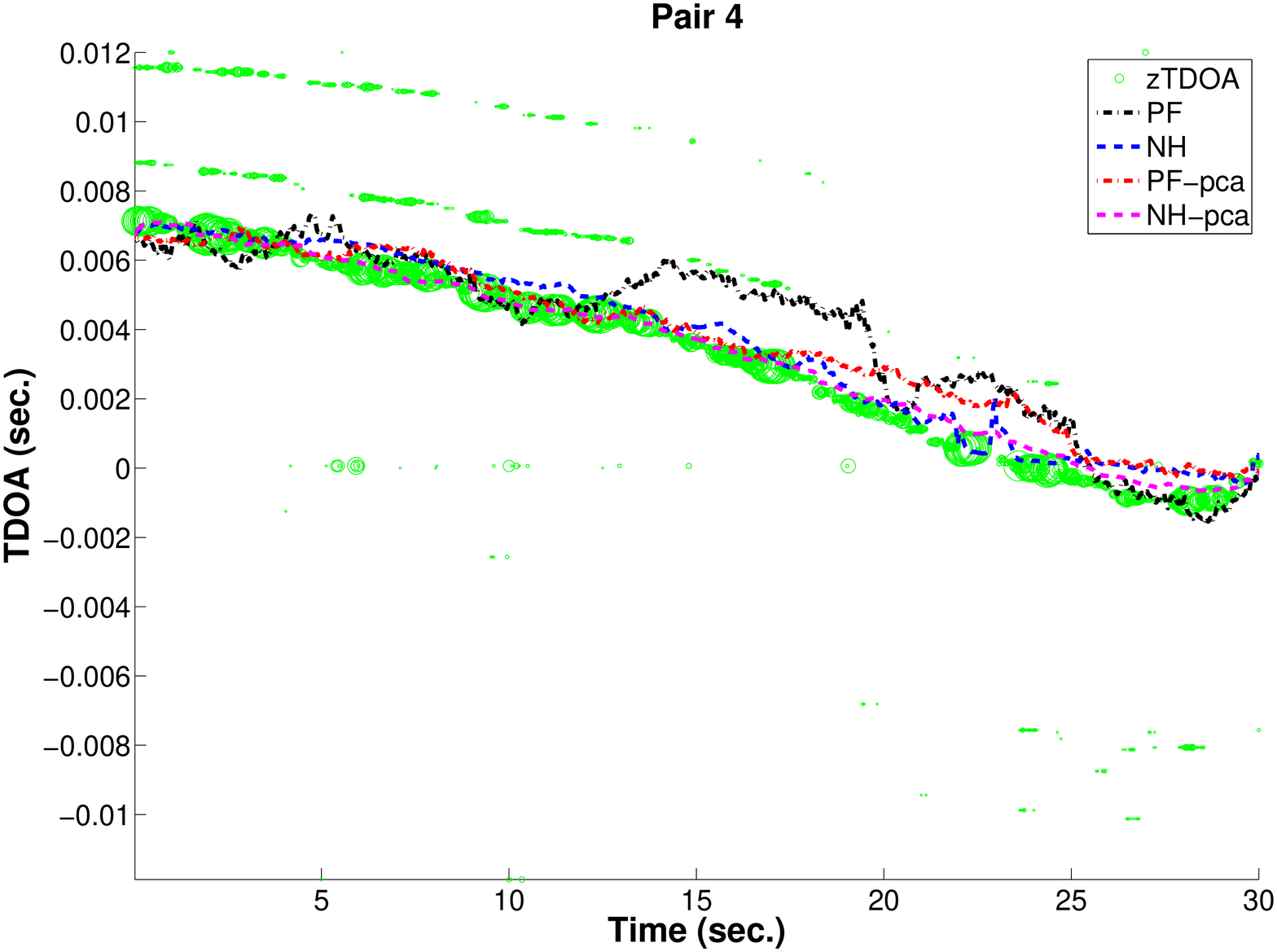}
\includegraphics[width=\columnwidth]{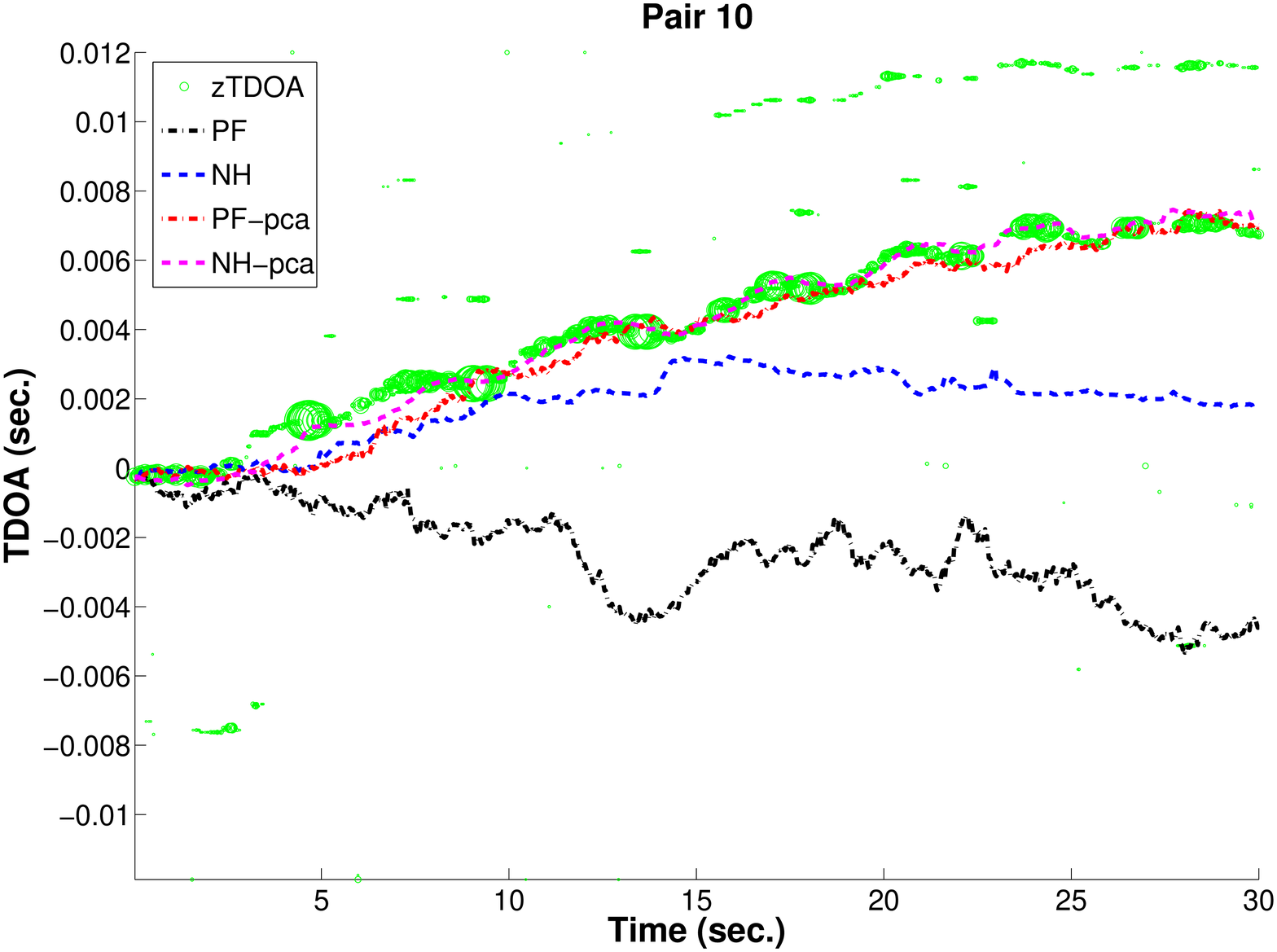}
}
\caption{Performance of NH and PF with and without using a global PCA projection for denoising.}
\label{fig:pca}
\end{center}
\vskip -0.2in
\end{figure*} 

\subsection{Integrating the Manifold Model}
\label{sec:treeusage}
The manifold modeling from Section~\ref{rptrees} is integrated into both particle filtering algorithms very easily after the resampling stage.  A final step is added after resampling a particle to denoise it so that it lies on the model of the manifold.  The leaf node in the PD-tree that corresponds to the state of the particle is found.  To denoise it, the particle is then projected onto the affine space spanned by the top k PCA components stored in this leaf.

In the experiments that follow we explore several manifold models:
\begin{enumerate}
  \item \textbf{No manifold modeling:} no projection is performed after the resampling step.
  \item \textbf{Fixed depth manifold modeling:}  We grow each PD-tree to a fixed depth and use the leaf nodes at this depth as the manifold model.
  \item \textbf{Randomized manifold modeling:} We grow the tree to a fixed depth and we examine the path from root to leaf node the particle takes in the PD-tree.  We then choose one of the nodes along this path uniformly at random to be the node which we use for the projection.  We hope this randomized model has the ability to adapt over time to which levels of the tree are currently best at modeling the position of the sound source being tracked.
\end{enumerate}

\section{Experiments}
\label{experiments}

\subsection{Experimental Setup}
Recordings were made at 16 kHz on a 7 microphone array that is part of an interactive display placed in a large public lobby.  The room is approximately 10m x 13m x 5m in size.  Four of the microphones are placed at the corners of the display which is mounted on one of the walls in the room, and the three remaining microphones are placed on the ceiling of the room.  For more details of the microphone setup and the room see~\cite{Cheamanunkul09, Ettinger08}.

To build a PD-tree we first collected a training set of TDOA vectors from our microphone array.  We accomplished this by moving a white noise producing sound source around the room near typical locations that sitting or standing people would be interacting with the display.  This resulted in approximately 20000 training TDOA vectors to which we built a PD-tree of depth 2.  In each node of the PD-tree we store the mean of the training data and the top k=3 principal directions.
 
Here are the parameter settings we use for the experiments that follow.  We use m=50 particles for each type of particle filter examined.  Our frame size is 500 ms with an overlap of 25 ms.  We set $\Sigma_r = \frac{4}{r}I_D$, where $r$ is the sampling rate.  The discounting factor for NH is set to $\alpha=0.05$, and the parameters of Equation~(\ref{eqn:scoring}) are $\sigma^2_z = 10$ and $Z_0 = 1$.

We made several real audio recordings of a person walking throughout the room facing the array and talking.  We describe each experiment in detail in what follows.

\begin{figure*}[t]
\vskip 0.2in
\begin{center}
\centerline{
\includegraphics[width=\columnwidth]{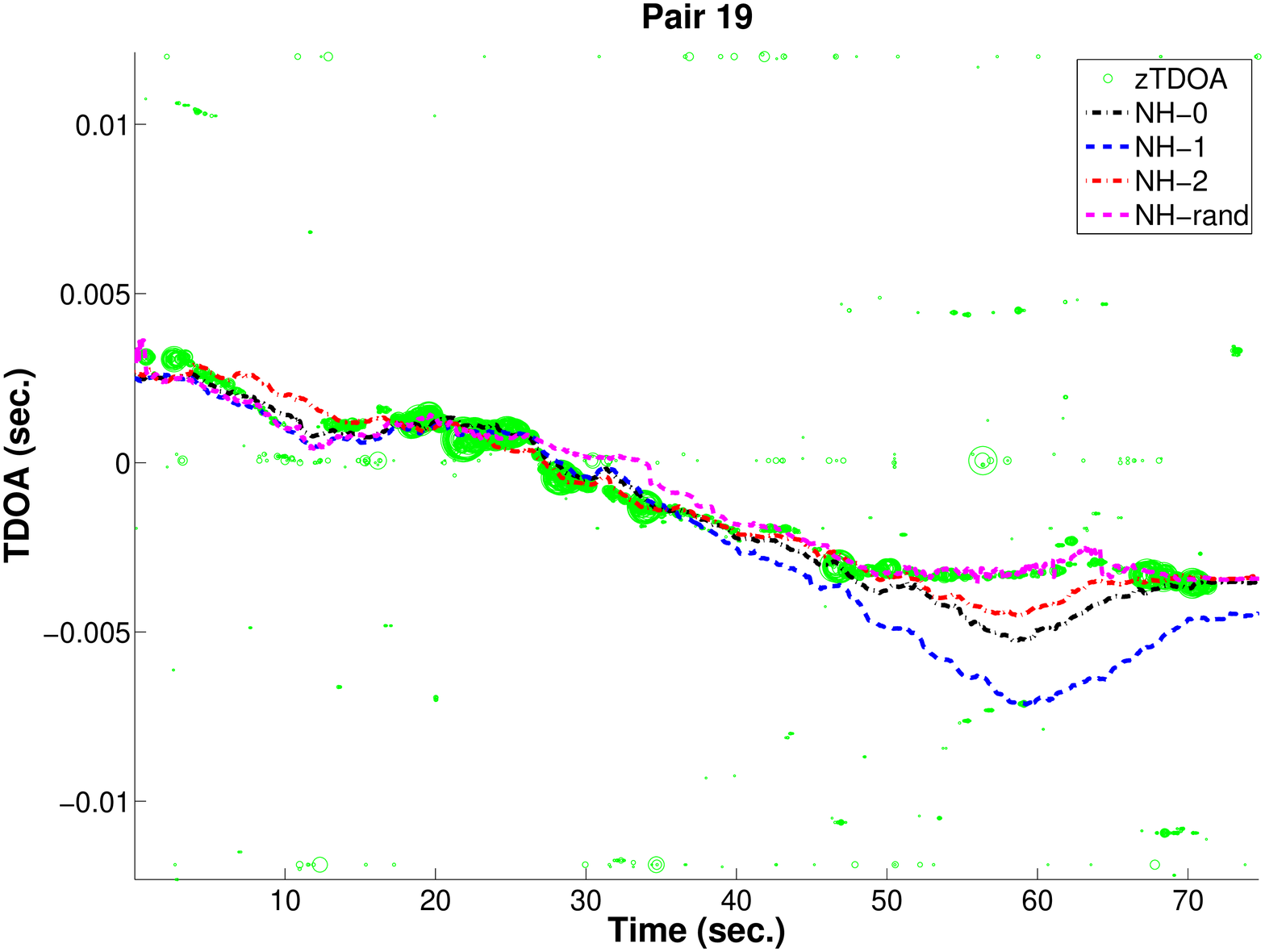}
\includegraphics[width=\columnwidth]{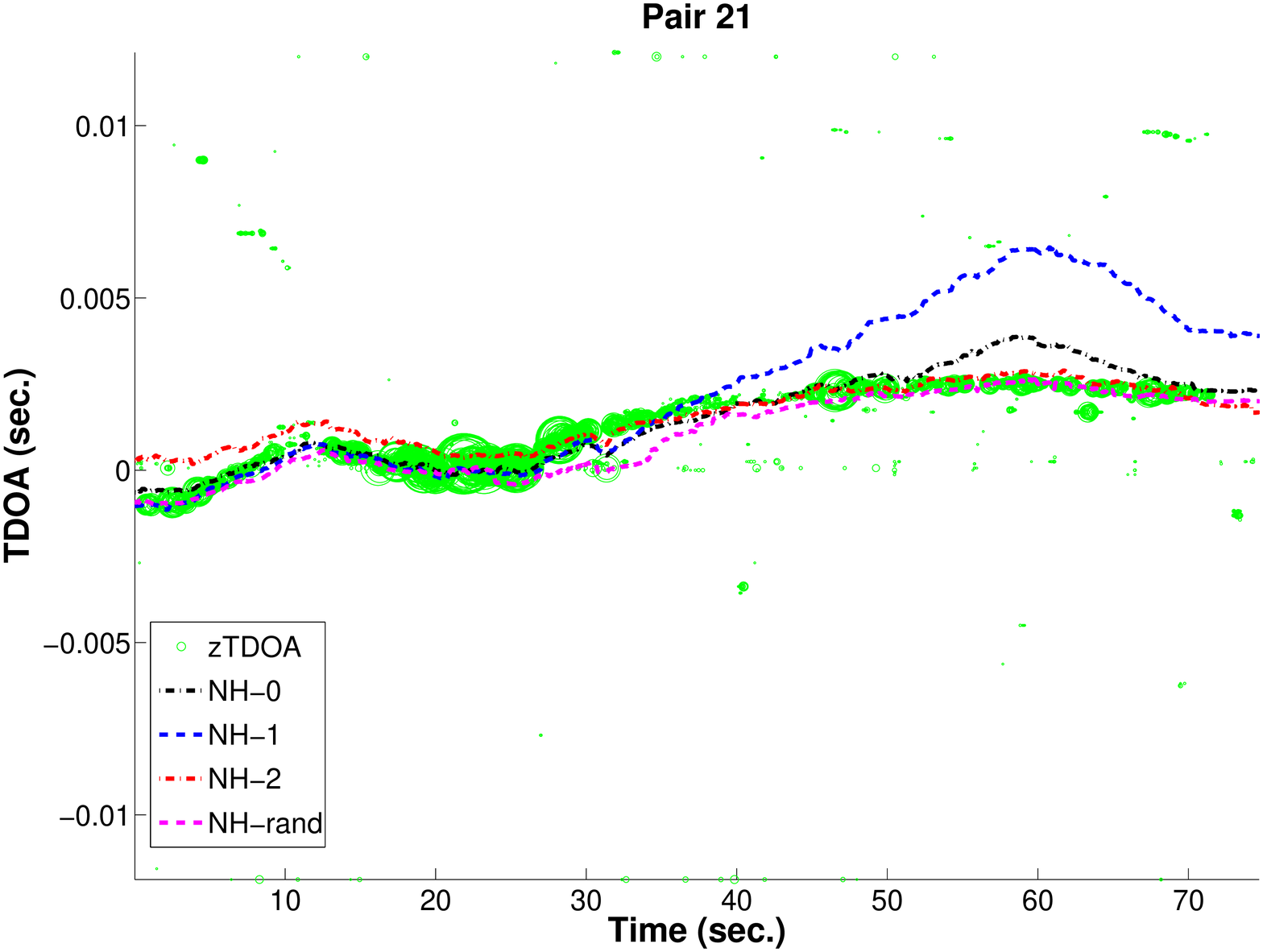}
}
\caption{Using various depths in the PD-tree as part of the projection step.}
\label{fig:depths}
\end{center}
\vskip -0.2in
\end{figure*} 

\subsection{Usage of Manifold Modeling}
This first experiment has a person walking and counting aloud while facing the array.  The person's path goes through the center of the room far from each microphone.  Since TDOAs evolve more slowly when the sound source is far from each microphone we'd expect this to be well modeled by the root PCA of our PD-tree.  Here we compare using the root PCA of our PD-tree versus no projection step at all for both SIR particle filters (PF) and the normal hedge particle filters (NH).  

Figure~\ref{fig:pca} depicts such a comparison.  Here we show tracking results from two microphone pairs that are typical of the remaining pairs.  In green is shown $Z_t^p$ where its magnitude is represented by the size of the circle marker.  The sound source moved in a continuous and slowly moving path so we'd expect each TDOA coordinate to follow a continuous and slowly changing path as well.  The trackers with the PCA projection step are able to follow the sound source, while the versions without the projection lose the source quickly.  

Remember that there are only 50 particles to track a state that is 21 dimensional.  There are no dynamics involved in our particle filters, so the resampling stage alone has to include enough randomness for the source to be tracked as it moves.  When the manifold model is not used the amount of randomness needed is to large for 50 particles to be able to track on all D dimensions.  However, when a model of the manifold is used effective tracking results can be had.  Moreover, it should be noted that the normal hedge version uses less randomness since it only resamples when the weight of a particle becomes zero.  Despite this, the normal hedge versions are able to have a competitive performance with SIR particle filters with much less randomness being used.

\subsection{Testing Different Manifold Models}
The setup of this experiment is exactly the same as the last except the path the speaker took traveled much closer to some pairs of microphones at certain points in time.  When a sound source is moving close to some set of microphones, the TDOAs involved with those microphones will change much more rapidly and in a much more non-linear way.  With this path we hope to examine the usefulness of deeper nodes in the PD-tree.  Since the performance of PF and NH are comparable when using the global PCA projection we only examine NH in this experiment.

Figure~\ref{fig:depths} is a similar figure to that discussed in the previous section.  The particle filtering variants examined here use projections at fixed depth zero (NH-0), one (NH-1), and two (NH-2).  The random strategy discussed in Section~\ref{sec:treeusage} is also examined (NH-rand).  It is clear that somewhere between 50-70s. the location of the sound source is modeled poorly by the global PCA at the root and is better modeled by the PCA at level 2.  However, it is only for this short duration where this modeling transition takes place.  Depth's 0 and 1 performed particularly poorly in this region, while depth 2 seems to have a significant advantage.

\begin{figure}[t]
\vskip 0.2in
\begin{center}
\centerline{
\includegraphics[width=\columnwidth]{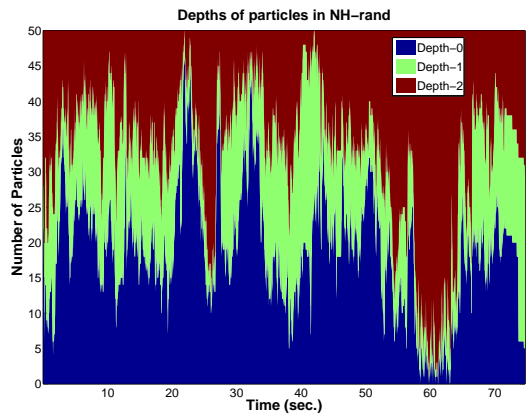}
}
\caption{For NH-rand, the PD-tree depths at time t that the m particles have been sampled from last.}
\label{fig:randbar}
\end{center}
\vskip -0.2in
\end{figure}

However, the best performing tracker was one that utilized the entire tree structure in a random fashion.  By allowing particles to die and birth randomly, there was a clear pressure to transition from a depth-0 model to a depth-2 model rather quickly by NH-rand.  This can be seen in Figure~\ref{fig:randbar}.  Here we depict what proportion of the 50 particles at time t were last sampled from which depth by a stacked bar graph.  There is a clear preference for transitioning towards depth-2 at this particular time period.  Nearly all the particles during this time period that were sampled from depth-2 are staying alive during this period.

This is a rather intuitive result since a particular node's PCA model may only be good for tracking in a small region of the entire 21 dimensional space that its PD-tree node represents.  When the sound source exits this region, some other depth in the tree may become a better model.  Using the randomness over time by NH-rand naturally captures such transitions.

\begin{figure}[t]
\vskip 0.2in
\begin{center}
\centerline{
\includegraphics[width=0.8\columnwidth]{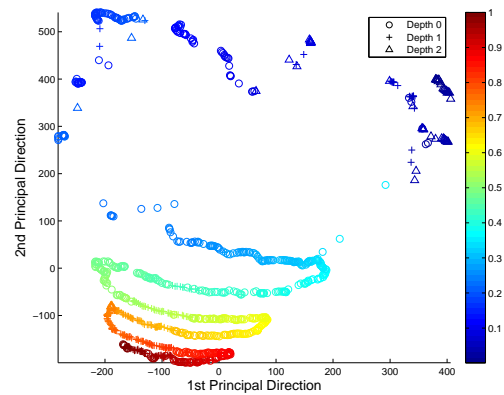}
}
\caption{Sweeping path for NH-rand on top 2 principal directions of root PCA.\label{fig:depth-mapping}}
\end{center}
\vskip -0.2in
\end{figure}

Figure~\ref{fig:depth-mapping} shows a sound source moving at constant speed a back-and-forth sweeping path.  Each sweep starts beyond one side of the display and continues across and past the opposite end of the display.  This is repeated at various distances away from the display.  The TDOA vectors predicted by NH-rand are projected on the top 2 principal components of the root PCA.  Colors indicate time, dark blue being the earliest part of the path that started approximately 1m from the display and red is the last segment of the path approximately 12m away.  The change in TDOAs is greatest when near the microphones on the display which results in a wide spacing of points.  The markers indicate which of the 3 depths the majority of the NH-rand particles were last sampled from.  In the center of the room its clear that the root-PCA performs best, whereas near the display on the right side depth 2 dominates, and far from the display depth 1 is best.

\section{Conclusion}
In this work we examine particle filtering methods for tracking the TDOA vectors for moving sound sources.  This is an essential problem to solve for audio localization and sound enhancement applications.  We present a model of the manifold based on space partitioning trees that alleviates the problem of high dimensional tracking with particle filters.  We also present a new version of a particle filter based on results from online learning that is competitive with traditional particle filters on this task and has properties that are attractive to many real world problems.  

\begin{small}
\bibliography{cameraman_uai2010}

\begin{thebibliography}{10}

\bibitem{arulampalam2002}
M.~Arulampalam, S.~Maskell, N.~Gordon, and T.~Clapp.
\newblock {A tutorial on particle filters for online nonlinear/non-Gaussian
  Bayesian tracking}.
\newblock {\em IEEE Transactions on signal processing}, 50(2):174--188, 2002.

\bibitem{Chaudhuri09}
K.~Chaudhuri, Y.~Freund, and D.~Hsu.
\newblock A parameter-free hedging algorithm.
\newblock In {\em Advances in Neural Information Processing Systems 22}, pages
  297--305. 2009.

\bibitem{Chaudhuri-2-09}
K.~Chaudhuri, Y.~Freund, and D.~Hsu.
\newblock Tracking using explanation-based modeling.
\newblock Technical report, UC San Diego, 2009.

\bibitem{Cheamanunkul09}
S.~Cheamanunkul, E.~Ettinger, M.~Jacobsen, P.~Lai, and Y.~Freund.
\newblock Detecting, tracking and interacting with people in a public space.
\newblock In {\em ICMI-MLMI '09: Proceedings of the 2009 international
  conference on Multimodal interfaces}, pages 79--86, 2009.

\bibitem{Dasgupta09}
S.~Dasgupta and Y.~Freund.
\newblock Random projection trees for vector quantization.
\newblock {\em Information Theory, IEEE Transactions on}, 55(7):3229--3242,
  July 2009.

\bibitem{Dibiase01}
J.~DiBiase, H.~Silverman, and M.~Brandstein.
\newblock Robust localization in reverberant rooms.
\newblock {\em In Microphone arrays: signal processing techniques and
  applications}, page 157, 2001.

\bibitem{Ettinger08}
E.~Ettinger and Y.~Freund.
\newblock Coordinate-free calibration of an acoustically driven camera pointing
  system.
\newblock In {\em Distributed Smart Cameras, 2008. ICDSC 2008. Second ACM/IEEE
  International Conference on}, pages 1--9, Sept. 2008.

\bibitem{Freund07}
Y.~Freund, S.~Dasgupta, M.~Kabra, and N.~Verma.
\newblock {Learning the structure of manifolds using random projections}.
\newblock In {\em Advances in Neural Information Processing Systems 20}. 2007.

\bibitem{Gordon93}
N.~Gordon, D.~Salmond, and A.~Smith.
\newblock {Novel approach to nonlinear/non-Gaussian Bayesian state estimation}.
\newblock {\em IEE proceedings. Part F. Radar and signal processing},
  140(2):107--113, 1993.

\bibitem{Samory09}
S.~Kpotufe.
\newblock Escaping the curse of dimensionality with a tree-based regressor.
\newblock In {\em COLT '09: Proceedings of the 22nd annual workshop on
  computational learning theory}, 2009.

\bibitem{Lehmann07}
E.~Lehmann and A.~Johansson.
\newblock Particle filter with integrated voice activity detection for acoustic
  source tracking.
\newblock {\em EURASIP J. Appl. Signal Process.}, 2007(1):28--28, 2007.

\bibitem{Verma09}
N.~Verma, S.~Kpotufe, and S.~Dasgupta.
\newblock Which spatial partition trees are adaptive to intrinsic dimension?
\newblock In {\em The 25th Conference on Uncertainty in Artificial
  Intelligence}. 2009.

\bibitem{Ward03}
D.~Ward, E.~Lehmann, and R.~Williamson.
\newblock Particle filtering algorithms for tracking an acoustic source in a
  reverberant environment.
\newblock {\em IEEE Transactions on Speech and Audio Processing}, 11:826--836,
  2003.

\end{thebibliography}
\bibliographystyle{abbrv}
\end{small}

\end{document}